\documentclass[12pt,a4paper,twoside]{article}
\usepackage{amsmath,amssymb}
\usepackage[english]{babel}
\usepackage{caption}
\usepackage{fancyhdr}
\usepackage[T1]{fontenc}
\usepackage{multirow}
\usepackage[pdftex]{graphicx}
\usepackage{hyperref}
\usepackage[all]{nowidow}
\usepackage{siunitx}
\usepackage{tikz}

\newcommand{\EatDot}[1]{}
\newcommand{\TheTitle}{\texorpdfstring{VideoKifu, or the automatic\\transcription of a Go game.}{VideoKifu, or the automatic transcription of a Go game.}}
\let\oldfootnote\footnote
\renewcommand\footnote[1]{\oldfootnote{\hspace{0.2em}#1}}
\usetikzlibrary{arrows.meta,backgrounds,fit,positioning,shapes.geometric}

\fancypagestyle{plain}{\fancyhead{}}
\hypersetup{
  pdftitle={\TheTitle},
  pdfauthor={Andrea Carta and Mario Corsolini},
  pdfsubject={\TheTitle},
  colorlinks=true,
  citecolor=[RGB]{0,80,0},
  linkcolor=[RGB]{120,0,0},
  urlcolor=[RGB]{0,0,160}}

\begin{document}

\phantomsection
\addcontentsline{toc}{section}{Front matter}
\author{Andrea Carta\thanks{~Both authors contributed equally.} \\ \url{http://www.micini.net/} \and Mario Corsolini\textsuperscript{$*$} \\ \url{http://www.oipaz.net/}}
\title{\TheTitle}
\date{}
\maketitle

\phantomsection
\addcontentsline{toc}{section}{Abstract}
\begin{abstract}
  In two previous papers~\cite{CC15,CC16} we described the techniques we employed for reconstructing the whole move sequence of a Go game. That task was at first accomplished by means of a series of photographs, manually shot, as explained during the scientific conference held within the LIX European Go Congress (Liberec, CZ). The photographs were subsequently replaced by a possibly unattended video live stream (provided by webcams, videocameras, smartphones and so on) or, were the live stream not available, by means of a pre-recorded video of the game itself, on condition that the goban and the stones were clearly visible more often than not.

  As we hinted in the latter paper, in the last two years we have improved both the algorithms employed for reconstructing the grid and detecting the stones, making extensive usage of the multicore capabilities offered by modern CPUs. Those capabilities prompted us to develop some asynchronous routines, capable of double-checking the position of the grid and the number and colour of any stone previously detected, in order to get rid of minor errors possibly occurred during the main analysis, and that may pass undetected especially in the course of an unattended live streaming. Those routines will be described in details, as they address some problems that are of general interest when reconstructing the move sequence, for example what to do when large movements of the whole goban occur (deliberate or not) and how to deal with captures of dead stones --- that could be wrongly detected and recorded as ``fresh'' moves if not promptly removed.
\end{abstract}

\section{Introduction}

Beyond the graphical user interface, VideoKifu\footnote{\url{http://www.oipaz.net/VideoKifu.html}} mainly consists of two continuously interacting engines: the grid tracking routines and the stone detection routines. Both engines are in turn split in two parts: a main one running as fast as possible in real time and a slower yet more accurate one running as a separate asynchronous thread.

When the recording of a game proceeds smoothly there is no need for the separate supervising routines, nevertheless during several real life experiments we realised that unforeseen events do occur, quite frequently indeed, sometimes causing situations that require a manual intervention to be solved. We added the supervising routines in order to minimise such manual interventions to a minimum (hopefully none). They slowly but constantly checks for inconsistencies between the location of the grid computed by the program and the actual one in the frames, as well as between the recognized stones and the actual ones. When a difference is detected for a prefixed number of consecutive frames, a warning is issued to the main routines that may use the info to amend the situation.

The program's structure is depicted in figure~\ref{fig:Flowchart}. Initialisations\footnote{Such as selecting video source, choosing rules and entering other info about the game.} and outputs\footnote{Such as SGF file, kifu, log files and HTML pages.} excepted, these are its main tasks:
\begin{figure}[p]
  \centering
  \begin{tikzpicture}
    \tikzstyle{base}=[draw,fill=yellow!10,minimum height=2em,semithick,text centered]
    \tikzstyle{decision}=[base,diamond,aspect=1.618,text width=4em]
    \tikzstyle{io}=[base,trapezium,trapezium left angle=75,trapezium right angle=105,text width=3em]
    \tikzstyle{process}=[base,text width=8em]
    \tikzstyle{startstop}=[base,rounded corners,text width=4em]
    \tikzstyle{back}=[inner xsep=2.5em,inner ysep=2.5em,rounded corners,text centered]
    \tikzstyle{backred}=[back,fill=red!20,inner xsep=1.5em,inner ysep=1.5em]
    \tikzstyle{backgreen}=[back,fill=green!20]
    \tikzstyle{backblue}=[back,fill=blue!20,inner xsep=0.5em]
    \tikzstyle{linebare}=[draw,semithick]
    \tikzstyle{line}=[linebare,-LaTeX]
    \tikzstyle{linedouble}=[line,LaTeX-LaTeX]
    \path node (start) [startstop] {Start};
    \path (start.east)+(2.0,0.0) node (input) [io] {Data\\input};
    \path (input.south)+(0.0,-0.965) node (init) [process] {Initial grid\\location};
    \path (init.south)+(0.0,-2.0) node (load) [io] {Load\\frame};
    \path (load.south)+(0.0,-1.0) node (filter) [process] {Frame filtering};
    \path (filter.west)+(-1.60,-1.0125) coordinate (dummy1);
    \path (filter.east)+(4.125,0.0) node (save) [io] {Data\\saving};
    \path (filter.south)+(0.0,-2.235) node (match) [process] {MatchTemplate};
    \path (filter.south)+(6.125,-2.235) node (hough) [process] {Linear Hough /\\Circular Hough};
    \node [below=of filter] (grid) {};
    \path (match.south)+(0.0,-2.42) node (main) [process] {Main cycle};
    \path (match.south)+(6.125,-2.42) node (super) [process] {Asynchronous\\cycle};
    \node [below=of match] (stones) {};
    \path (super.north)+(0.0,1.35) node (dummy2) [text width=12em,semithick] {};
    \path (main.south)+(0.0,-2.34) node (next) [decision] {Another\\frame?};
    \path (next.west)+(-2.09,0.0) coordinate (dummy3);
    \path (next.south)+(0.0,-2.035) node (output) [io] {Data\\output};
    \path (output.east)+(2.0,0.0) node (end) [startstop] {End};
    \path [line] (start) -- node {} (input);
    \path [line] (input) -- node {} (init);
    \path [line] (init) -- node {} (load);
    \path [line] (load) -- node {} (filter);
    \path [line,dashed] (filter) -- node {} (save);
    \path [line] (filter) -- node {} (match);
    \path [linedouble,dashed] (hough) -- node {} (match);
    \path [linedouble] (match) -- node {} (main);
    \path [linebare,dashed] (main) -| node {} (dummy1);
    \path [line,dashed] (dummy1) -| node {} (save);
    \path [linedouble,dashed] (super) -- node {} (main);
    \path [line] (main) -- node {} (next);
    \path [linebare] (next.west)+(0.02,0.0) -- node [above right] {Yes} (dummy3);
    \path [line] (dummy3) |- node {} (load);
    \path [line] (next.south)+(0.0,0.01) -- node [above right] {No} (output);
    \path [line] (output.east)+(-0.005,0.0) -- node {} (end);
    \begin{scope}[on background layer]
      \node (mp) [backgreen] [fit=(load) (filter) (match) (main) (next)] {\rule{1.042em}{0pt}\textbf{Main process}\\\rule{0pt}{30.632em}};
      \node (ap) [backblue] [fit=(dummy2) (save) (hough) (super)] {\textbf{Asynchronous processes}\\\rule{0pt}{18.94em}};
      \node (grid) [backred] [fit=(match) (hough)] {\textbf{Grid tracking}\\\rule{0pt}{3.676em}};
      \node (stones) [backred] [fit=(main) (super)] {\textbf{Stones detection}\\\rule{0pt}{3.818em}};
    \end{scope}
  \end{tikzpicture}
  \caption{flowchart of VideoKifu's operation.}
  \label{fig:Flowchart}
\end{figure}
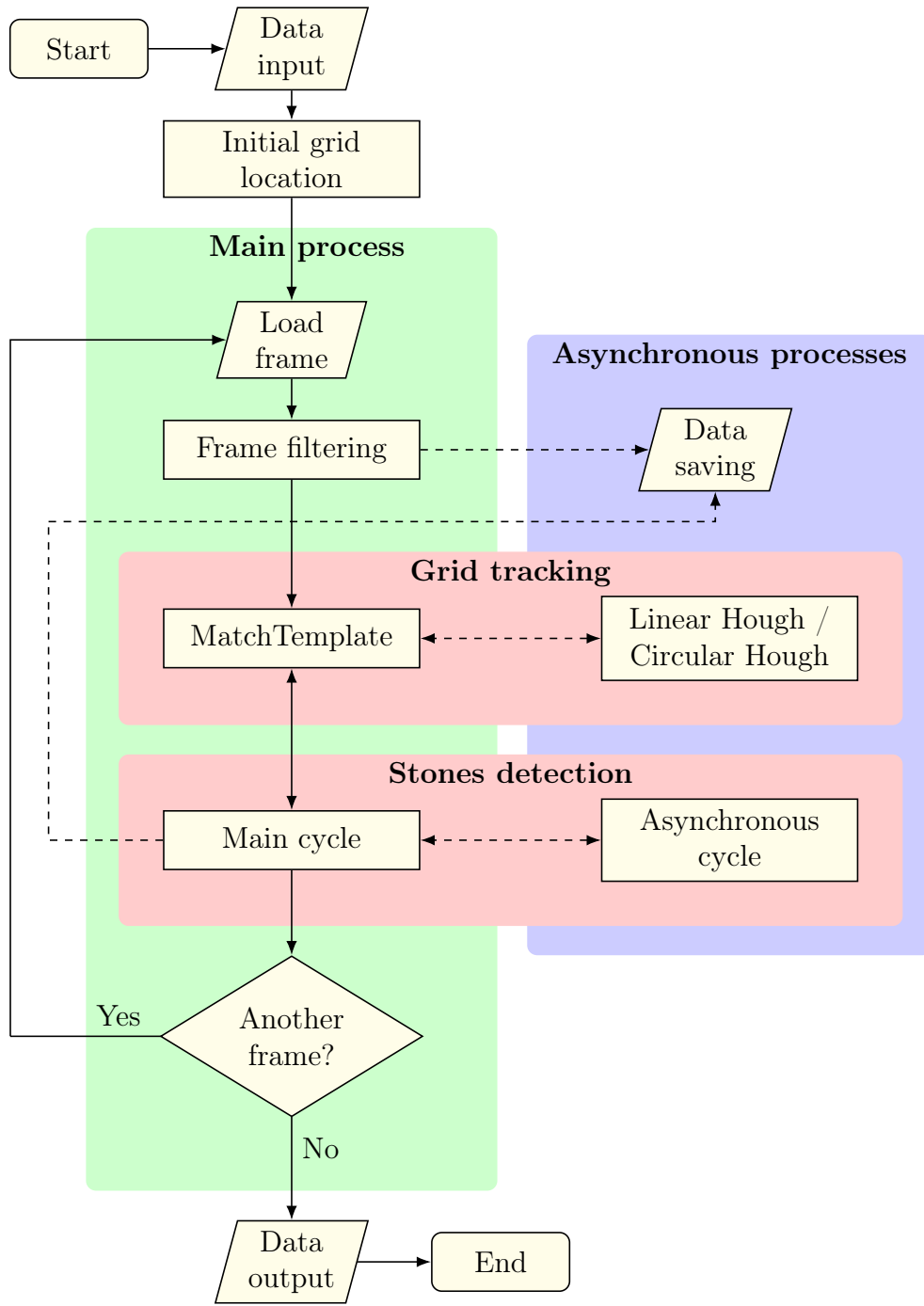

\begin{itemize}
  \item Initial grid location.\\Based on linear Hough transform, the details of the algorithm are expounded in~\cite[\S2.1]{CC16}. Since then it received only minor improvements, mostly regarding how to counter shifts of the grid caused by a border of the goban wrongly recognised as a grid line.
  \item Frame acquisition and filtering.\\Mostly carried out by means of OpenCV\footnote{\url{http://opencv.org/}} functions.
  \item Grid tracking.\\This task is implemented by three different routines. A fast one based on the MatchTemplate function of OpenCV, always running as a main process and described in~\cite[\S2.2]{CC16}. Since it is able to track only small movements of the grid, there are two more powerful routines running in separate threads whose purpose is to track large movements: the former is the same used for the initial grid location, usable in the first part of a game (when grid lines are mostly visible); the latter is a routine based on the circular Hough transform, described in section~\ref{sec:Grid}, usable in the last part of a game (when lots of stones are visible).
  \item Stones detection.\\This task too is split in two parts: one always running as a main process, whose basic functioning is described in~\cite[\S3]{CC15}, while various specific issues are addressed in~\cite[\S3]{CC16} (modifications needed to deal with a video stream instead of a series of photographs) and in section~\ref{sec:Captures} (where the tricky problem of dealing with captures is tackled). The second part of the task is the one running as a separate asynchronous thread; it is a boosted version of the main routine and it is described in section~\ref{sec:Stones}.
\end{itemize}

\section{Improving grid's tracking}
\label{sec:Grid}

In~\cite[\S2.1]{CC16} we described the algorithm used at the beginning of an analysis to identify the initial location of the goban (or, to be more precise, of the grid of lines painted on it), based on a double application of the linear Hough transform. As expected, experiments in the field confirmed that such algorithm maintains a good rate of success even when many stones have been played, concealing large parts of the grid: it generally succeeds provided that at least half of the points on the goban are empty and visible. Furthermore, it executes fast enough\footnote{About 0.3 seconds for a clean 1080p frame on our usual (not so new) personal computer of reference, equipped with an Intel$^\circledR$ Core{\texttrademark} i7-4770 CPU @ 3.40GHz and integrated HD Graphics 4600.} to be used as a superintending separate task to double check the accuracy of the recognised grid.

That check is necessary if we want to permit unattended analysis, as the automatic micro-recalibration (described in~\cite[\S2.2]{CC16}) is able to follow movements of the grid not larger than about one stone's radius between consecutive frames. That is enough to manage vibrations and light bumps on the goban or on the table or on the camera, but it would be useless in case of a strong bump or any other kind of major disturbance affecting the framing. Furthermore, another issue arose during games with plenty of moves: in order to counteract unavoidable long-term drift phenomena, we evaluate a linear regression of the pixels on the external line of the grid; that works well if a large part of the external lines are visible, but it could fail when those lines are covered by lots of stones. In that case the algorithm strongly relies on the small parts of the lines left free, so it could get confused by player's hands or arms or even heads, if their edges appear in the frame in front of the points the program believes to be empty (and thus usable for evaluating the aforesaid linear regression). In those situations even the edges of a stone not yet recognized may alter the outcome of the recalibration.

So we modified the linear Hough routine in order to use it for asynchronous checks as well. It is not super-fast, yet it is quite usable: it usually completes its computation while the main routines analyse about three frames. To avoid accidental errors a new grid is not accepted unless it remains stable for at least two or three consecutive frames (depending on circumstances and on the amount of points covered by stones). So, should the framing really be affected by some disturbance, the adjustment is usually computed within nearly two seconds. In most situation this is not a problem, even when the game has a fast pace, as the program is able to detect multiple stones in a single frame, once the correct location of the grid has been restored; otherwise we rely on the techniques expounded in section~\ref{sec:Stones}.

The real hindrance is that unfortunately it is not possible to use the linear Hough routine to check the correctness of the grid for an entire game: when the grid lines are mostly concealed it simply fails, so it is not usable in the final part of medium to long games (the so called yose). This is particularly harmful as the final moments of a Go game are usually the most agitated ones, with lots of moves played in a fast pace and with both players paying more attention to the clock than to the table or the tripod holding the camera: thus an accuracy check of the grid is particularly needed.

It is theoretically possible to use circular or elliptical Hough transform in order to reconstruct the location of the grid starting from the position of the known stones, either in the original frame (in which stones appear elliptical due to perspective effects) or in a rectified one (in which they are almost circular, provided the perspective has not substantially changed from the last correct grid). The circular Hough transform usually requires the same amount of time to be calculated as the linear one, while the elliptical transform is quite slower, especially because it could not be always possible to predict the eccentricity of the ellipses to be searched. That happens because the camera recording the game is usually put on the side of the goban, so the farthest stones are seen from a lower angle of view: in such a situation the shape that stones project onto the camera plane is deeply influenced by their thickness, which may substantially vary among different stone sets (moreover they may even be flat on one side). Even the shape projected in the rectified plane may vary, but in this case at least a good portion of the outline of the stones is almost circular, allowing the circular Hough transform to detect them, provided the search parameters are not too strict.

Circular transform is thus preferable and we exploited it to devise a new algorithm able to pinpoint the location of the grid in yose.

\subsection{Use of circular Hough transform}

\begin{figure}[!hbt]
  \frame{\includegraphics[width=0.542857\linewidth]{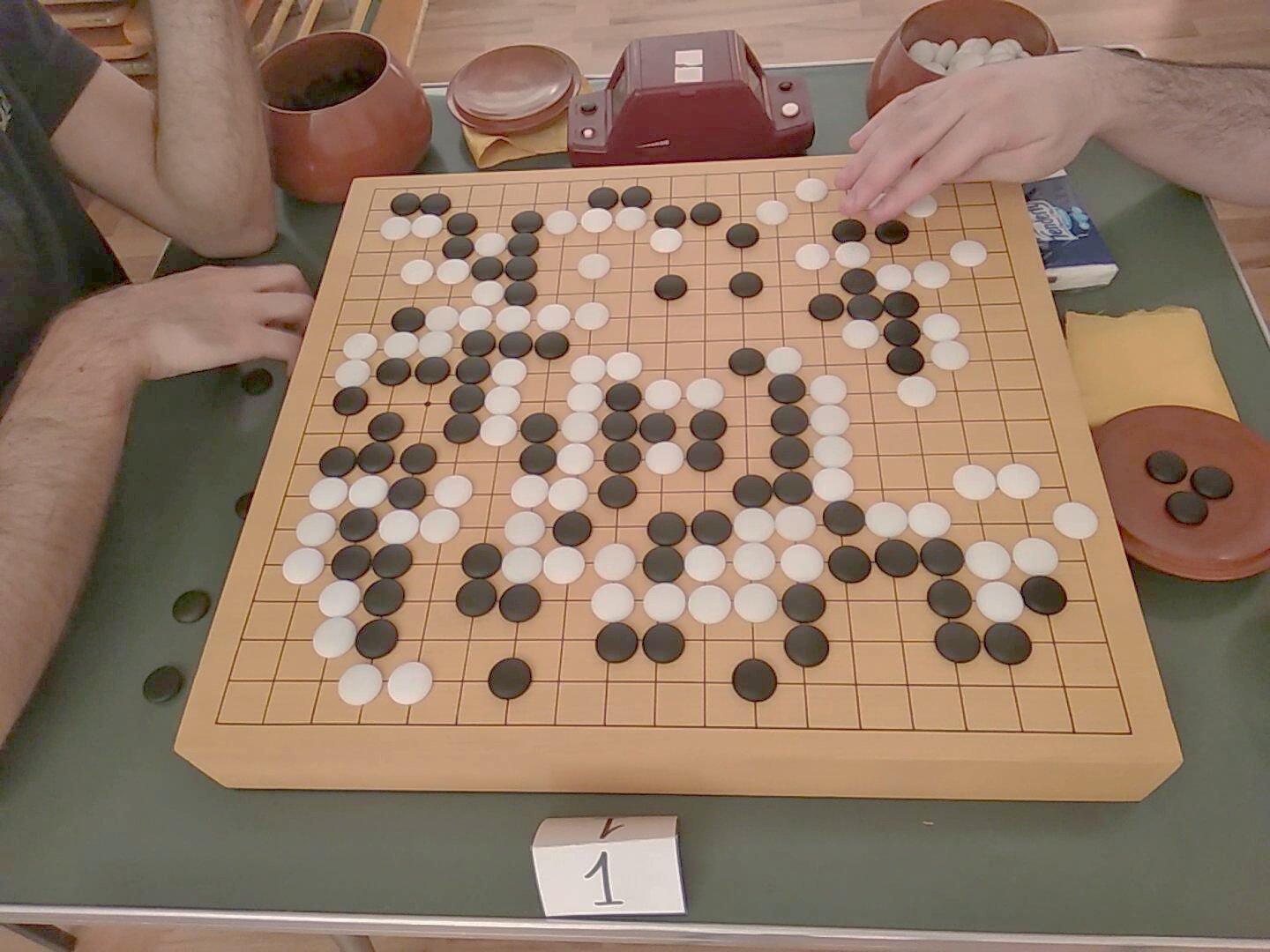}}
  \hfill
  \frame{\includegraphics[width=0.407143\linewidth]{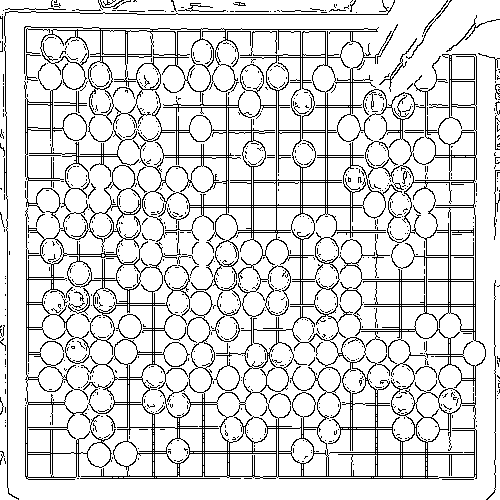}}
  \caption{the final position of Carlo Metta [4d] vs. Alessandro Martinelli [1d], Bologna Tournament 2018, IV round. Original frame ($1440 \times 1080$ pixels) on the left, filtered rectification ($500 \times 500$ pixels) on the right.}
  \label{fig:Metta-Martinelli}
\end{figure}

First of all we create a small rectified image of the part in the new frame covered by the last known grid. The reason it has to be small\footnote{Compared to the original frames captured from the camera.} can be seen in figure~\ref{fig:Metta-Martinelli}, showing one of the final frames of the Metta-Martinelli game in Bologna Tournament 2018. Even with a good (high) angle of view, rectifying an angled image like the one depicted on the left of the figure means to stretch its upper side more than its lower side. Stretching too much would blur the borders of the stones, making them invisible to the edge detector (Canny filter, in this case).

\begin{figure}[!hbt]
  \begin{minipage}{0.475\linewidth}
    \frame{\includegraphics[width=\linewidth]{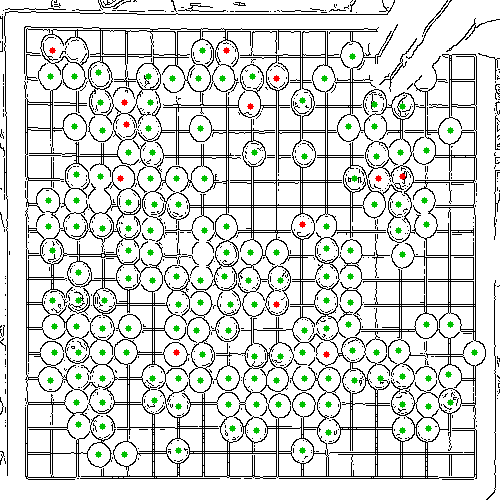}}
    \caption{centres of the circles found in figure~\ref{fig:Metta-Martinelli} (right).}
    \label{fig:FilteredCentres}
  \end{minipage}
  \hfill
  \begin{minipage}{0.475\linewidth}
    \frame{\includegraphics[width=\linewidth]{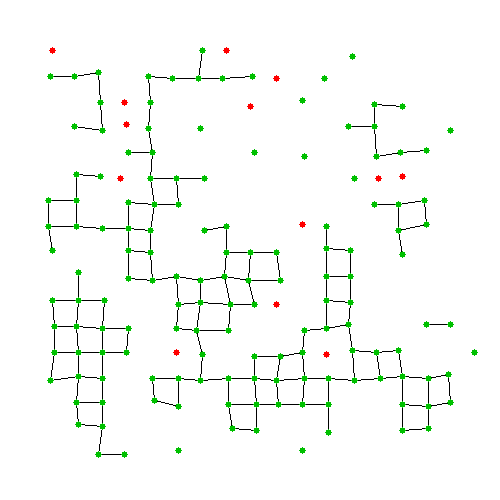}}
    \caption{segments joining adjacent stones.}
    \label{fig:Links}
  \end{minipage}
\end{figure}

Circular Hough transform is then applied to the rectified image. Projected stones should maintain an almost circular shape even when the goban has been moved among frames, provided the perspective does not change too much (a condition easily met as the goban usually lies on a table: its incidental movements should be coplanar --- and rigid). In spite of that, usually the circles found by the Hough transform are not always the same, even between contiguous frames, and they are not always exactly located in the same spot. The reason may vary: stones' shadows on the goban, reflections (especially on black stones), stones too tightly packed with overlapping borders, changes in light conditions\dots they all contribute to alter the output of the circular Hough transform. Besides, Canny filter renders white stones slightly bigger than black ones, adding another element of uncertainty in the transform. For all those reasons, we deem as real stones only the circles that appear in almost the same spot for at least two consecutive frames. In figure~\ref{fig:FilteredCentres} red dots represent the centres of newly recognised circles, green dots are those recognised for at least two consecutive frames (positional tolerance is the radius of each dot); as shown, in that frame a few stones were not recognised at all.

Another step that must be taken is the one depicted in figure~\ref{fig:Links}: the slopes of all the segments joining adjacent confirmed stones are evaluated and their mean value is used to amend incidental rotations of the goban (figure~\ref{fig:Links}, for instance, needs to be rotated by $\ang{0.4}$ clockwise).

Once the centres are correctly rotated, their best match with known stones is searched for. Movements of up to a third of goban's side between frames are managed if at least about a half of the stones\footnote{To be more precise, the minimum common area could be as low as $\frac{4}{9}$ of the goban.} are in common between the previous known position of the grid and the actual one in the frame under analysis.

Once the pairing has been established between the centres of each recognised circle and the coordinates of actual known stones, it is possible to evaluate where the corners of the grid should be in the rectified image (even if they are external) and to back-project them into the original frame. Accuracy is of paramount importance: as discussed, the rectified image is smaller than the original frame, so, for instance, in case of a 1080p video source, an error of one pixel in the former is back-projected into an error of about two pixels in the latter.

\begin{figure}[!hbt]
  \begin{minipage}{0.407143\linewidth}
    \frame{\includegraphics[width=\linewidth]{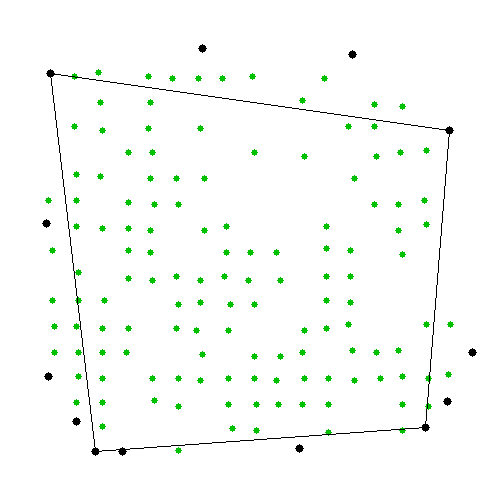}}
    \caption{external centres and quadrilateral of maximum area.}
    \label{fig:MaxArea}
  \end{minipage}
  \hfill
  \begin{minipage}{0.542857\linewidth}
    \frame{\includegraphics[width=\linewidth]{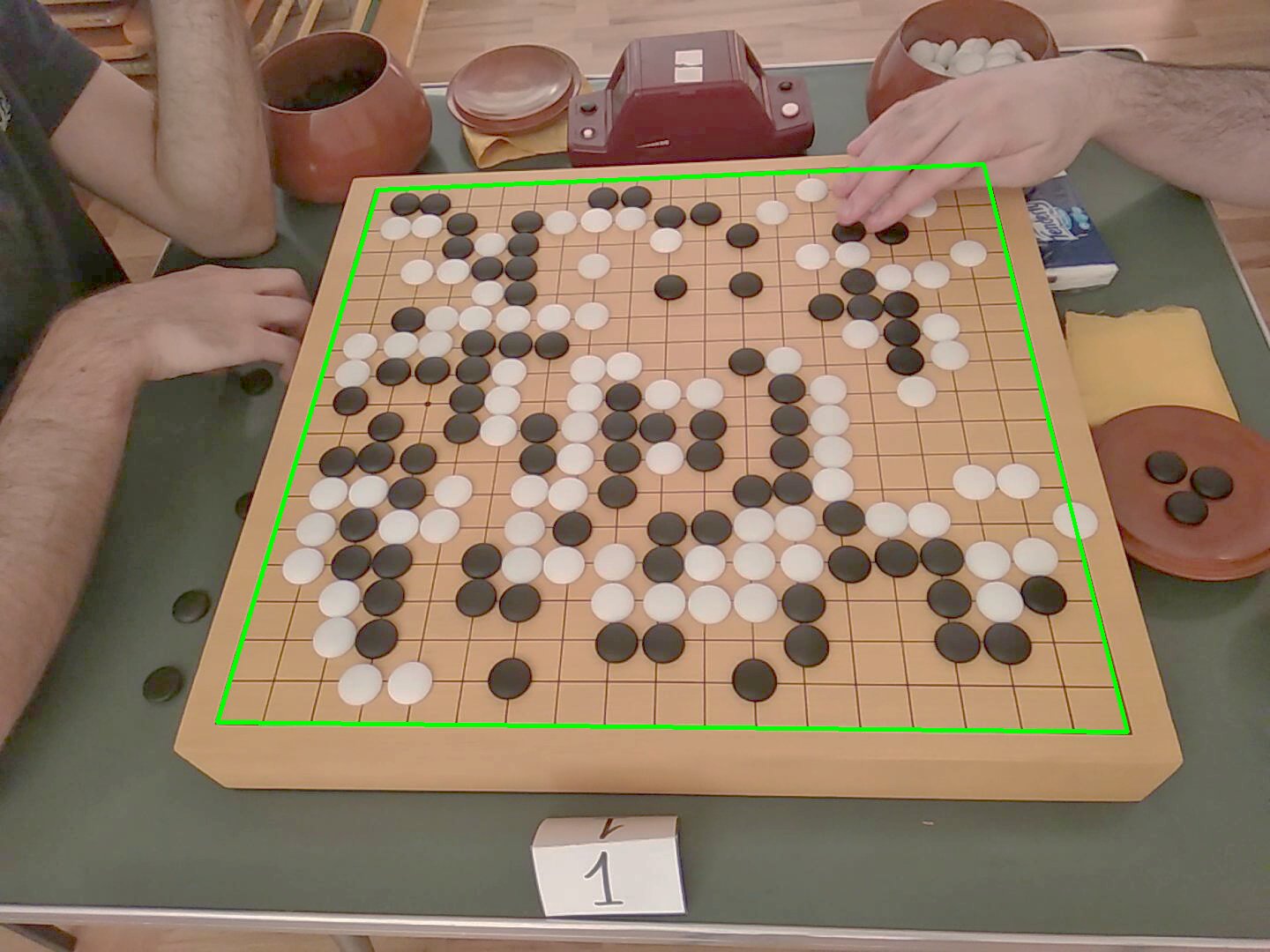}}
    \caption{back-projection (in green) of the evaluated grid.}
    \label{fig:BackProjection}
  \end{minipage}
\end{figure}

Experiments showed that using all the centres to compute the best fitting back-projecting matrix leads to inaccurate results, while selecting four centres produces better and stables ones. Anyway, as the stones are often misplaced and their locations are not exactly found by the circular Hough transform (figure~\ref{fig:FilteredCentres} is a piece of evidence), using only four centres is also a source of potential inaccuracy, so we evaluate the weighted mean\footnote{Weights are the normalised areas of each quadrilateral.} of the back-projection matrix based on the four centres that enclose the maximum area and up to seven other matrices based on as many other quadrilaterals, whose vertices are randomly chosen from the external centres\footnote{That is the vertices of the convex hull of all the centres.} (as shown in figure~\ref{fig:MaxArea}).

The resulting location of the grid is usually quite accurate, as shown in figure~\ref{fig:BackProjection}, with accuracy increasing with the number of stones placed on the goban. Actually, so accurate that we chose to switch from the algorithm based on linear Hough transform to this one well before the linear algorithm commences to fail. The computation is also very fast, as the transform is applied to an image much smaller\footnote{A black and white image of 0.25 Mpixels at most.} than the original frames: it usually runs in 30 ms or less.

\section{Detecting errors in the placement of stones}
\label{sec:Stones}

In~\cite[\S3.3.3]{CC15} we described the algorithm used for detecting the stones: its main idea was to compare the features (luminance, chrominance and so on) of the stones we were looking for with the ones' already detected on the goban. Similar features (computed by means of an elaborate function) meant the point under scrutiny was likely covered by a stone, and vice versa. This algorithm is almost flawless and insensitive to typical disturbance (hands passing over the goban and such), but under extreme circumstances may fail. For example, stones completely out of centre are difficult to detect, and if the light is pale (such as inside a room when the sky is cloudy) the goban looks grey and the chrominance feature becomes useless. This, in turn, means the function is weakened and becomes more sensible to other kind of errors (stones out of centre, disturbance, strong reflections): this does not prevent the stones to be detected but several seconds may pass before that,\!\footnote{In most cases it takes three frames to detect a stone, that is about half to one second.} and in many circumstances (joseki and yose, for example) other stones may be played in the while; if too many undetected stones are on the goban the algorithm may fail and the analysis cannot continue.

To prevent such an unpleasant occurrence, a new algorithm has been implemented: constantly running in a separate asynchronous process it scans the whole goban looking for stones, closely examining all promising points, and not only the ones that strongly resemble a stone; this way it is possible to detect ``dubious'' stones that had been discarded by the main algorithm for some of the aforementioned reasons and insert them, albeit lately, in the move list (a warning will advice the user to check the order). If the pace is not too fast, the stone may likely be detected before any other has been played, and will be inserted in the correct place; but even if that were not the case the analysis could restart without further errors (for example, when a stone is missed the next ones are detected in the wrong order, as the algorithm expects the wrong colour), something that otherwise could never happen. The asynchronous algorithm is so powerful that it is capable to detect even dozens of missing stones in a matter of seconds, bringing back to life an analysis that would never have been restarted. It becomes even possible to start an analysis when the game has already begun, something we thought unsuitable before, partly because we weren't able to detect more than four stones at the same time, partly because such moves cannot be detected in the correct order, a situation that takes the user some time to be fixed.

Of course the algorithm should never fail because, should a not-existent stone be detected, this would lead to disaster; that's why the presence of a stone is double and tripled checked, in a much more accurate way than the main algorithm's. Five frames instead of the usual three are needed to deem the stone as real; and each time both the usual function and the Hough transform are employed in order to check that. This means that the algorithm is slow, and cannot detect stones as fast as the main one; but, given the fact they are not detected in the correct order and must be checked in any case after the game, the issue does not matter.

Also, the algorithm scans for false positives, trying to detect and eliminate them; but in this case the matter is much more delicate: there could be many reasons why a stone is falsely detected (one of the most frequent will be explained in the next section) and usually it has been played some time before, so deleting it could worse matters. In such cases the stone is not deleted but a warning is issued; the only case in which the stone is deleted is when it had been the last move played, as in this case, and this case only, no damage may be done because of its deletion (an example of such a case will also be discussed in the next section).

\section{Dealing with captures}
\label{sec:Captures}

In~\cite[\S3.3]{CC16} we described what looked like the best way to handle the killing of several stones at once. That way proved to be inadequate, for reasons we will now explain, while describing exhaustively the problem and its solution.

When a capture occurs, VideoKifu must detect not only the killing stone, but also the ones removed from the goban, and update its image accordingly. If only a stone has been captured, the task is trivial. But the situation drastically changes when several ones are captured at once.
In such a case, there are two possible ways to proceed:
\begin{itemize}
  \item when the killing stone is detected, all the captured stones are immediately deleted from goban's image and VideoKifu begins waiting for the next move;
  \item when the killing stone is detected, VideoKifu waits for all the captured stones to be removed, and only after that begins waiting for the next move.
\end{itemize}

Both ways are flawed. In the first way, if the captured stones are not removed at once VideoKifu could --- and likely will --- detect these stones and believe they have just been put on the goban: because their colour is now expected to move, one of them will be mistaken for the next stone played and, unless manually deleted, its presence will corrupt the whole move list, up to the end of the game.
That's why we thought the second way was the best one, and wrote so in~\cite[\S3.3]{CC16}, but eventually realised that was not true. It happened that during a game White captured three black stones, and Black played at once a snap-back move, killing in turn one white stone. The snap-back move was played before VideoKifu had realized some black stones were missing,\!\footnote{In~\cite[\S3.3]{CC16} it was pointed out that VideoKifu's purpose is detecting stones, not the opposite: detecting empty points instead is difficult and requires some time.} and as one of them had reappeared, the program never stopped waiting for it to eventually disappear, thus never restarting the analysis.

It is quite obvious that such an occurrence must be avoided at all costs, so we went back to the first way to handle captures, trying to find a way that could prevent stones not immediately removed to be mistaken for new moves. The asynchronous process was crucial in accomplishing that, because it can detect false positives, that are empty points on the goban mistaken for stones; also, if the false positive coincides with the move last played it is deleted at once, and this is exactly what we needed. It is easy to understand the reason: as we explained before, when the captured stones are not immediately removed from the goban one of them will possibly be detected, mistaken for the new move and added to the move list; but once eventually removed it will become a false positive and the asynchronous process will delete it from the move list, as indeed it's the last move played. If, in the same situation, a snap-back is played --- as we saw happen --- the stone will be detected all the same, but not being removed any time soon the asynchronous process will do nothing and the move will correctly remain in the move list. Kos, when stones are repeatedly played and removed, are not a concern, as only one stone is involved and cannot be put again on the goban before other moves are played.

The whole process can be summarized as follows:
\begin{itemize}
  \item when the killing stone is detected, all the captured stones are removed from the program's goban's image at once and VideoKifu begins waiting for the next move;
  \item if the stones are not removed from the real goban at once VideoKifu will again detect one of them, wrongly assuming it is the expected move;
  \item when this stone is eventually removed (something that must happen before any other move is played) the asynchronous process will acknowledge it as a false positive and will delete it from the move list, restoring the correct situation;
  \item should the stone be played again, this time for real, before being deleted from the move list, the asynchronous process will do nothing;
  \item should a different move be played before VideoKifu could delete the wrong one from the move list, this time the main routine will see that the stone last played has disappeared and will replace the last move with the real one.
\end{itemize}

Captures of many stones at once are not common, but occur in most games nonetheless; the process we have just described should hopefully handle all kind of situations that could arise from such an event.

\section{Conclusion}

At the moment --- July 2018 --- we have 18 games available, one of them recorded by two different cameras (a DSLR in video mode and a tablet). The outcomes of the analyses are shown in table~\ref{table:Videos} (in the ``Frames'' column the upper numbers indicate the frames saved in the video, the lower ones are the actual ones processed by VideoKifu; in live analyses the program saves just the frames actually analysed, and that's why the two real/processed values are identical).
\begin{table}[p]
  \centering
  \small
  \begin{tabular}{|r|l||c|c|r|l|}
    \hline
    \multirow{2}{*}{\#} & Game:                       &\multirow{2}{*}{Moves}&   Resolution   & Frames                &Manual\\
                        & place (notes)               &                      &   device       &re/pr\rule{.423em}{0pt}&interventions\\
    \hline
    \hline
    \multirow{2}{*}{ 1.}& Carta-Corsolini:            & \multirow{2}{*}{96}  & 640$\times$480 &                24,844 &\multirow{2}{*}{none}\\
                        & friendly game (13$\times$13)&                      &   DSLR         &                 6,211 &\\
    \hline
    \multirow{2}{*}{ 2.}& Carta-Corsolini:            & \multirow{2}{*}{96}  &1920$\times$1080&                29,795 &\multirow{2}{*}{none}\\
                        & friendly game (13$\times$13)&                      &   tablet       &                 5,959 &\\
    \hline
    \multirow{2}{*}{ 3.}& Pignelli-Albano:            & \multirow{2}{*}{233} &1440$\times$1080&               268,017 &\multirow{2}{*}{none}\\
                        & Pisa 2015                   &                      &   tablet       &                67,004 &\\
    \hline
    \multirow{2}{*}{ 4.}& Pantalone-Balzaretti:       & \multirow{2}{*}{143} &1920$\times$1080&\multirow{2}{*}{16,740}&\multirow{2}{*}{none}\\
                        & Pisa 2016 (live)            &                      &   webcam       &                       &\\
    \hline
    \multirow{2}{*}{ 5.}& De Lazzari-Greenberg:       & \multirow{2}{*}{231} & 640$\times$480 &               128,093 &\multirow{2}{*}{none}\\
                        & Pisa 2016                   &                      &   smartphone   &                24,336 &\\
    \hline
    \multirow{2}{*}{ 6.}& Ragno-Gioia:                & \multirow{2}{*}{190} & 640$\times$480 &                66,947 &\multirow{2}{*}{none}\\
                        & Pisa 2016                   &                      &   smartphone   &                22,316 &\\
    \hline
    \multirow{2}{*}{ 7.}& Telesca-Metta:              & \multirow{2}{*}{260} &1920$\times$1080&                78,274 &\multirow{2}{*}{none}\\
                        & Pisa 2016                   &                      &   tablet       &                15,654 &\\
    \hline
    \multirow{2}{*}{ 8.}& Martinelli-van den          & \multirow{2}{*}{262} &1280$\times$720 &               100,469 &\multirow{2}{*}{none}\\
                        & Busken: Roma 2016           &                      &   smartphone   &                33,490 &\\
    \hline
    \multirow{2}{*}{ 9.}& Potort\`{i}-De Lazzari:     & \multirow{2}{*}{242} &1920$\times$1080& \multirow{2}{*}{5,434}&many (bad fo-\\
                        & Pisa 2018 (live)            &                      &   webcam       &                       &cus and light)\\
    \hline
    \multirow{2}{*}{10.}& Nunziati-Potort\`{i}:       & \multirow{2}{*}{199} &1600$\times$1080&               116,736 &many (bad fo-\\
                        & Pisa 2018                   &                      &   webcam       &                23,347 &cus and light)\\
    \hline
    \multirow{2}{*}{11.}& Piccinno-Sanzone:           & \multirow{2}{*}{174} &1600$\times$1080&               154,341 &none (with\\
                        & Pisa 2018                   &                      &   webcam       &                30,868 &USM filtering)\\
    \hline
    \multirow{2}{*}{12.}& Spallanzani-Piccinno:       & \multirow{2}{*}{271} &1920$\times$1080&\multirow{2}{*}{16,164}&none (with\\
                        & Pisa 2018 (live)            &                      &   webcam       &                       &USM filtering)\\
    \hline
    \multirow{2}{*}{13.}& Mieli-Fanti:                & \multirow{2}{*}{139} & 960$\times$720 &\multirow{2}{*}{16,071}&\multirow{2}{*}{none}\\
                        & Roma 2018 (live)            &                      &   smartphone   &                       &\\
    \hline
    \multirow{2}{*}{14.}& Ragno-Fanti:                & \multirow{2}{*}{113} & 960$\times$720 &\multirow{2}{*}{20,756}&\multirow{2}{*}{none}\\
                        & Roma 2018 (live)            &                      &   smartphone   &                       &\\
    \hline
    \multirow{2}{*}{15.}& Martinelli-Forte:           & \multirow{2}{*}{240} &1920$\times$1080&               166,290 &\multirow{2}{*}{none}\\
                        & Roma 2018                   &                      &   tablet       &                33,258 &\\
    \hline
    \multirow{2}{*}{16.}& Martinelli-Parton:          & \multirow{2}{*}{252} &1920$\times$1080&                79,497 &\multirow{2}{*}{none}\\
                        & Roma 2018                   &                      &   tablet       &                26,499 &\\
    \hline
    \multirow{2}{*}{17.}& Hueber-Fanti:               & \multirow{2}{*}{207} & 960$\times$720 &\multirow{2}{*}{24,502}&none in the\\
                        & Bologna 2018 (live)         &                      &   smartphone   &                       &program\\
    \hline
    \multirow{2}{*}{18.}& Metta-Sgaravatti:           & \multirow{2}{*}{128} &1440$\times$1080& \multirow{2}{*}{1,637}&\multirow{2}{*}{none}\\
                        & Bologna 2018 (live)         &                      &   smartphone   &                       &\\
    \hline
    \multirow{2}{*}{19.}& Metta-Martinelli:           & \multirow{2}{*}{166} &1440$\times$1080&\multirow{2}{*}{27,347}&\multirow{2}{*}{none}\\
                        & Bologna 2018 (live)         &                      &   smartphone   &                       &\\
    \hline
  \end{tabular}
  \normalsize
  \caption{analysed videos, available online at VideoKifu's homepage.}
  \label{table:Videos}
\end{table}

It was only at Pisa 2018\footnote{\url{http://www.europeangodatabase.eu/EGD/Tournament_Card.php?&key=T180310D}} that we stumbled upon a serious problem: there was no room for a tabletop tripod and we found out our Logitech C615 webcam cannot correctly focus past one meter. It was therefore difficult to detect and track the grid (many of its lines were blurred) and also many times the stones were detected later than usual. In the end we could only analyse the Spallanzani-Piccinno and the Piccinno-Sanzone games without manual interventions (applying an appropriate ``UnSharp Mask'' filtering), because they were played under a good light, while the two ones played by Francesco Potort\`{i}, that were more blurred and played under a worse light, require too many manual interventions before reaching the end. We stumbled upon a similar problem at Pisa 2016,\!\footnote{\url{http://www.europeangodatabase.eu/EGD/Tournament_Card.php?&key=T161022B}} when we recorded two games at 480p by means of a smartphone, and the low resolution, paired with the small screen, prevented us to correctly focus the goban; this makes the grid difficult to detect (USM filter helps) but, once detected, no further problems occur and the analysis never require manual intervention. A peculiar case is the game between Niccol\`{o} Sgaravatti and Carlo Metta, played at Bologna 2018:\footnote{\url{http://www.europeangodatabase.eu/EGD/Tournament_Card.php?&key=T180609B}} it was recorded by means of an inadequate, rented notebook, equipped with an old two (instead of the required four) core CPU, and although no problems were encountered, the analysis was so slow that a serious lag soon manifested and, once in yose, grew up to two minutes; therefore too many moves were detected at the same time, of course in random order, and eventually we decided to stop the recording after 128 moves (each one correctly detected), believing the live analysis had become futile by then. Another peculiar case was the Hueber-Fanti game of the same tournament: it was necessary to slightly move the bowl of white stones as it was in such a position that one stone inside the bowl appeared in the frames as if it was over the goban.

All the other games present no problems at all: the grid is detected at once, correctly tracked throughout the game, and all the stones are detected in the correct order (with a few exceptions when, usually in yose, three or more stones are detected at the same time, something the user is always warned of). Also, when the analysis were live, each one of them reached the end of the game without problems, except for the aforementioned ones at Pisa 2018.

Of course, further improvements are still desirable; for example we are looking for ways to improve the frames' pre-filtering in order to reduce the possible blurring and the effect of bad light (we found out that artificial light is preferable to natural one, as it enhances the colour's difference between the white/black stones and the yellowish/ochre goban). We are also trying to further improve the grid tracking routine, which is bound to the precision of the HoughLines/HoughCircles OpenCV functions, which in turn is not always reliable as it should be expected.

At the moment, if the hardware employed is good (a four core CPU is mandatory, as well as good quality cameras, with manual focus and no distortions, something that unfortunately rules out most webcams), VideoKifu v1.1.0 should be able to correctly analyse any game, both live and deferred, without errors or manual interventions.

\bibliography{References}
\bibliographystyle{alpha}
\addcontentsline{toc}{section}{References}

\end{document}